\documentclass[conference,10pt,a4paper]{IEEEtran}
\IEEEoverridecommandlockouts

\usepackage{silence}
\usepackage{amsmath,amssymb,amsfonts,mathtools}
\usepackage{amsthm}
\usepackage{bm}
\allowdisplaybreaks

\usepackage{graphicx}
\usepackage{booktabs}
\usepackage{xcolor}
\usepackage{array,multirow}
\usepackage{tabularx}
\usepackage{makecell}
\usepackage{adjustbox}
\usepackage{xcolor}
\usepackage[hypcap=false]{caption}
\usepackage{subcaption}
\usepackage{float}
\usepackage{longtable}
\usepackage[table]{xcolor}
\usepackage{colortbl}
\usepackage{microtype}
\usepackage{enumitem}
\setlist[itemize]{itemsep=2pt,topsep=2pt,leftmargin=*}
\setlist[enumerate]{itemsep=2pt,topsep=2pt,leftmargin=*}

\usepackage{siunitx}
\newcolumntype{N}{S[table-format=+1.6]}
\newcolumntype{L}[1]{>{\raggedright\arraybackslash}p{#1}}

\usepackage{algorithm}
\usepackage{algpseudocode}

\usepackage[
    colorlinks=true,
    citecolor=blue,
    linkcolor=blue,
    urlcolor=blue
]{hyperref}
\usepackage[nameinlink,noabbrev]{cleveref}
\crefformat{equation}{[Eq.~(#2#1#3)]}
\crefformat{figure}{[Fig.~#2#1#3]}
\crefformat{table}{[Tab.~#2#1#3]}
\crefformat{section}{[Sec.~#2#1#3]}
\crefformat{subsection}{[Sec.~#2#1#3]}
\crefformat{theorem}{[Thm.~#2#1#3]}
\crefformat{lemma}{[Lem.~#2#1#3]}
\crefformat{proposition}{[Prop.~#2#1#3]}
\crefformat{corollary}{[Cor.~#2#1#3]}
\crefformat{definition}{[Def.~#2#1#3]}
\crefformat{remark}{[Rem.~#2#1#3]}
\crefformat{example}{[Ex.~#2#1#3]}

\theoremstyle{plain}
\newtheorem{theorem}{Theorem}

\newtheorem{proposition}[theorem]{Proposition}

\theoremstyle{definition}
\newtheorem{definition}[theorem]{Definition}

\theoremstyle{remark}

\newcommand{\qedbox}{\hfill$\square$}

\usepackage{cite}

\newcommand{\feat}[2]{\ensuremath{#1}\,\textnormal{(\emph{#2})}}
\newcommand{\Xone}{\feat{X_1}{Loan duration}}
\newcommand{\Xtwo}{\feat{X_2}{Loan amount}}

\newcommand{\one}{\mathbf{1}}
\newcommand{\zero}{\mathbf{0}}
\newcommand{\mvec}{\mathbf{m}}

\DeclareMathOperator{\E}{\mathbb{E}}

\newcommand{\PP}{\mathbb{P}}
\newcommand{\vx}{\mathbf{x}}
\newcommand{\vX}{\mathbf{X}}

\begin{document}
\title{Aumann-SHAP: The Geometry of Counterfactual Interaction Explanations in Machine Learning}

\author{
\IEEEauthorblockN{Adam Belahcen}
\IEEEauthorblockA{\textsc{Airess} \& \textsc{Um6p}\\
adam.belahcen@um6p.ma}
\and
\IEEEauthorblockN{St\'ephane Mussard}
\IEEEauthorblockA{\textsc{Chrome}, Univ. N\^{\i}mes;\\
\textsc{Airess} \& \textsc{Um6p}\\
stephane.mussard@unimes.fr}
}

\maketitle

\begin{abstract}
We introduce \texttt{Aumann-SHAP}, an interaction-aware framework
that decomposes counterfactual transitions by restricting the model
to a local hypercube connecting baseline and counterfactual features.
Each hypercube is discretized into a grid to construct an induced
micro-player cooperative game in which elementary grid-step moves
become players. Shapley and LES values on this TU-micro-game yield
geometry-aware within-pot attributions that converge to the diagonal
Aumann--Shapley / Integrated Gradients limit under grid refinement,
and recover equal-split Shapley as the degenerate $m=1$ special case.
An exact grid-state closed form gives polynomial-time computation for
fixed interaction order. On a synthetic benchmark with known ground
truth, equal-split Shapley carries an irreducible bias while
Aumann-SHAP converges to the correct decomposition. On German Credit,
interaction geometry changes feature priority rankings in $12.3\%$ of
instances. On UCI Heart Disease, equal-split misattributes a
cholesterol suppressor as a positive contributor, which is a sign error
Aumann-SHAP corrects. On MNIST, game-theoretic attribution reaches
target confidence with $3.5\times$ fewer edits than magnitude-based
ordering, with micro-game Shapley achieving the best efficiency
across all budgets.
\end{abstract}


\section{Introduction}\label{sec:intro}

Machine learning models must be explainable, contestable, and actionable \cite{debock2024xaior}, which has motivated the development of explainable artificial intelligence (XAI). A dominant approach to XAI is feature attribution, where given an instance $\mathbf{x}$ and a predictor $g(\cdot)$, the
objective is to decompose a score $g(\mathbf{x})$ into feature-wise contributions that satisfy an \textit{efficiency}
property (the contributions sum to what is being explained).

The Shapley value \cite{shapley1953} provides an
axiomatic foundation for such decompositions and underlies widely used tools such as SHAP \cite{lundberg2017shap},
with applications spanning from human resources analytics to bank
performance evaluation \cite{shi2025bankdea} among others.
Beyond the classical Shapley value, related rules include Aumann--Shapley / Integrated Gradients for path-based settings \cite{aumannshapley1974,sundararajan2017ig} and interaction-aware indices such as Shapley--Taylor \cite{sundararajan2019shapleytaylor}. Different coalition-value definitions yield different explanation semantics \cite{sundararajan2020many}.
More generally, different modeling choices (\textit{e.g}., baselines and reference distributions) induce different ``Shapley values for explanation'' \cite{sundararajan2020many}. Finally, fair and efficient non-Shapley alternatives within
the Linear--Efficient--Symmetric (LES) family have also been advocated for attribution \cite{Condevaux2022}.

In many applications, however, the core question is not why the model predicted $g(\mathbf{x}^0)$ at a baseline point, but
how to change inputs to obtain a desired or feasible outcome $g(\mathbf{x}^1)$. This is the setting of counterfactual explanations and
algorithmic recourse \cite{wachter2017counterfactual,ustun2019actionable,mothilal2020dice,laugel2017inverse}.
Standard
generators include optimization-based formulations \cite{wachter2017counterfactual}, randomized search procedures
such as DiCE \cite{mothilal2020dice}, and boundary-search methods like Growing Spheres \cite{laugel2017inverse}.
While these methods identify which features change between $\mathbf{x}^0$ and $\mathbf{x}^1$, they typically do
not quantify how much each changed feature contributed to the improvement, nor do they explain how \textit{interactions} between features drive the
transition.\footnote{This gap is especially acute in nonlinear models (\textit{e.g}., ensembles), where a feature may be weak in
isolation but decisive in combination, and where effects can depend on the order/context of changes along the
baseline-to-counterfactual transition, motivating path-based interpretability analyses \cite{luborgonovo2024}.
Related concerns are also emphasized in the recourse literature, as nearest counterfactuals can fail to correspond to
actionable interventions \cite{karimi2021recourse,verma2020cfreview}.}

A natural way to explain counterfactual transitions is to attribute the \emph{difference} $\Delta y := g(\mathbf{x}^1)-g(\mathbf{x}^0)$ \textit{via} cooperative-game decompositions. Using Harsanyi dividends \cite{harsanyi1963}, it is possible to allocate interaction pots between features to each individual feature. The standard Shapley rule splits each pot equally among its members \cite{luborgonovo2024}, providing a symmetric baseline as a natural starting point for explainability of counterfactual paths. However, equal splitting ignores the geometry of the transition, treating all features symmetrically regardless of whether a feature enables an interaction early by entering a synergistic regime or harvests it only after other coordinates shifted.

\textit{Running example.} A loan applicant has baseline $\vx^0$ with no checking account ($X_8=0$) and poor credit history ($X_{13}=1$). The counterfactual $\vx^1$ flips both. The score improvement $\Delta V$ contains a pure interaction pot $\phi_{\{X_8,X_{13}\}}$, the synergy of having a checking account \emph{and} clearing credit history. Equal-split Shapley assigns half to each, blind to geometry. But if the interaction only activates once $X_8$ moves first, $X_8$ deserves more credit; it \emph{enables} the interaction. Aumann-SHAP detects this from the residual surface $r_u(t)$ on the local cube, assigning $X_8$ a larger share (Table~\ref{tab:within_top_realloc} shows a $7.4\%$ reallocation of $\Delta V$).

The main contributions are the following:
\begin{enumerate}
    \item We introduce a micro-player Transferable-utility game (TU-game) for counterfactual transitions by discretizing the local baseline counterfactual hypercube and treating grid-step moves as players.
    \item We derive the Aumann-SHAP (micro-game Shapley), an attribution method that yields geometry-aware, within-pot contributions of pure interactions between features (recovering the equal-split Shapley \cite{luborgonovo2024} as a special case).
    \item We generalize the framework to an Aumann--LES family, enabling robustness comparisons across attribution rules on the same counterfactual change.
    \item We provide an exact grid-state closed form that avoids enumerating $2^{n}$ micro-coalitions and gives polynomial-time computation in grid resolution (for fixed interaction order).
\end{enumerate}

Section~\ref{sec:section2} reviews related work on attribution and counterfactual explanations. Section~\ref{sec:section3} introduces the local-cube residual interactions and the induced micro-game construction. Section~\ref{sec:Aumann-SHAP}
presents micro-game-Shapley and Aumann-LES attribution methods and their theoretical properties. Section~\ref{sec:experiments} reports simulations and experiments. Section~\ref{sec:conclusion} concludes.

\section{State of the art}\label{sec:section2}

\subsection{Explainability and feature attribution}

XAI methods, \textit{i.e.} attribution methods for black-box models, fall into three families, and each one of them can potentially mask \textit{interactions} between features \cite{debock2024xaior}. The first relies on local surrogates that approximate $g(\cdot)$ around an input, as in LIME \cite{ribeiro2016lime}. A leading approach is Shapley attribution, which treats features as cooperative-game players and derives attributions from coalition values \cite{shapley1953,lundberg2017shap}. Shapley methods satisfy axioms such as symmetry and efficiency and work well for tabular data \cite{shapley1953,lundberg2017shap}, but different definitions of coalition values (\textit{e.g}., missingness, conditioning, causality) yield different explanations \cite{sundararajan2020many,frye2020asymmetric,heskes2020causal}. Alternative rules within the Linear--Efficient--Symmetric (LES) family trade off axioms and computation and allow robustness comparisons \cite{Condevaux2022}. The second family includes gradient and path-based methods for differentiable models \cite{debock2024xaior}. Integrated Gradients (IG) attributes prediction changes along a baseline-to-input path \cite{sundararajan2017ig} and connects to the continuous Aumann--Shapley value \cite{aumannshapley1974}. Vision-oriented saliency methods (\textit{e.g}., Grad-CAM, SmoothGrad) are common \cite{selvaraju2017gradcam,smilkov2017smoothgrad}, though our focus is on tabular data where interaction accounting is central. The third family uses perturbation and global sensitivity analysis, such as permutation importance, PDP, ICE, ALE, and Sobol indices \cite{breiman2001rf,friedman2001pdp,goldstein2015ice,apley2020ale,sobol1993sens}. While useful for global insight, they do not provide a single local additive decomposition for a baseline-to-target change, nor isolate pure interaction pot features.

\subsection{Explainability and counterfactual analysis}
Counterfactual explanations address a different question than pointwise attribution: instead of asking why
$g(\mathbf{x}^0)$ occurred (\textit{e.g.} why a client is not creditworthy), the goal is to find how inputs should change to reach a desired outcome $g(\mathbf{x}^1)$ (the client becomes creditworthy).

Much of the literature focuses on generating counterfactuals (\emph{generators}), such as
DiCE \cite{mothilal2020dice} and boundary-expansion strategies like Growing Spheres \cite{laugel2017inverse}.
Counterfactuals are typically non-unique: many feasible $\mathbf{x}^1$ may exist with similar distance but
different changed features, so robustness to the generator and its hyperparameters is important when explanations
guide actions or compliance narratives. While generators provide what changed between $\mathbf{x}^0$ and $\mathbf{x}^1$, they usually do not quantify how much
each change contributed to the shift $g(\mathbf{x}^1)-g(\mathbf{x}^0)$, nor how interactions mediated the shift.
This is especially problematic in nonlinear models, where effects are context- and path-dependent, motivating
path-based counterfactual explanation \cite{luborgonovo2024}.

\textit{Interaction attribution} within counterfactual analysis is essential for non-additive models. A common route is to define a cooperative game using
hybrid inputs and decompose the total change into Harsanyi interaction pots \cite{harsanyi1963}. In this context, the Shapley value \cite{shapley1953} allocates each pot by equal split among its members, yielding a Shapley-grounded baseline for
counterfactual transitions \cite{lundberg2017shap,luborgonovo2024}. However, the equal split issued from Shapley \cite{luborgonovo2024} is blind to
geometry, it cannot distinguish whether a feature enables an interaction early along the move from $\mathbf{x}^0$
to $\mathbf{x}^1$ or harvests it later after other coordinates have changed.

\textit{The problem.} Equal-split counterfactual Shapley \cite{luborgonovo2024} splits each Harsanyi pot $\phi_u$ equally among members of $u$, ignoring geometry. Consider $g(\mathbf x)=x_1x_2+x_1+x_2+1$ with $\mathbf x^0=(0,0)\to\mathbf x^1=(1,1)$. The transition of the interaction term $x_1x_2$, \textit{i.e.} the pot $\phi_{\{1,2\}}=1$, is split into $0.5$ for $x_1$ and $0.5$ for $x_2$ when both coordinates go from 0 to 1, \textit{ignoring the different transitions between 0 and 1}. Our contribution bridges counterfactual path geometry and attribution by restricting the model to the local hypercube, which contains all possible transitions between $\mathbf{x}^0$
and $\mathbf{x}^1$. 

Table~\ref{tab:method_comparison} positions \texttt{Aumann-SHAP}
against the closest related methods.

\begin{table}[H]
\centering
\caption{
\scriptsize\textbf{Comparison of attribution methods on key
properties. CF = designed for counterfactual transitions;
Geo = geometry-aware within-pot split; Int = interaction
pot decomposition; IG lim = converges to Aumann--Shapley/IG
limit; Poly = polynomial closed-form computation.}}
\label{tab:method_comparison}
\footnotesize
\setlength{\tabcolsep}{3pt}
\renewcommand{\arraystretch}{1.2}
\begin{tabular}{lccccc}
\toprule
\textbf{Method} & \textbf{CF} & \textbf{Geo} &
\textbf{Int} & \textbf{IG lim} & \textbf{Poly}\\
\midrule
SHAP \cite{lundberg2017shap}
  & \textcolor{red}{$\times$} & \textcolor{red}{$\times$}
  & \textcolor{red}{$\times$} & \textcolor{red}{$\times$}
  & \textcolor{red}{$\times$}\\
Integrated Gradients \cite{sundararajan2017ig}
  & \textcolor{red}{$\times$} & \textcolor{olive}{$\sim$}
  & \textcolor{red}{$\times$} & \textcolor{green!60!black}{$\checkmark$}
  & \textcolor{green!60!black}{$\checkmark$}\\
Shapley--Taylor \cite{sundararajan2019shapleytaylor}
  & \textcolor{red}{$\times$} & \textcolor{red}{$\times$}
  & \textcolor{green!60!black}{$\checkmark$} & \textcolor{red}{$\times$}
  & \textcolor{red}{$\times$}\\
Equal-split CF Shapley \cite{luborgonovo2024}
  & \textcolor{green!60!black}{$\checkmark$} & \textcolor{red}{$\times$}
  & \textcolor{green!60!black}{$\checkmark$} & \textcolor{red}{$\times$}
  & \textcolor{green!60!black}{$\checkmark$}\\
\texttt{Aumann-SHAP}
  & \textcolor{green!60!black}{$\checkmark$} & \textcolor{green!60!black}{$\checkmark$}
  & \textcolor{green!60!black}{$\checkmark$} & \textcolor{green!60!black}{$\checkmark$}
  & \textcolor{green!60!black}{$\checkmark$}\\
\bottomrule
\end{tabular}
\end{table}

Integrated Gradients attributes $\Delta y$
at the feature level but does not decompose interaction pots.
Shapley--Taylor explains
$g(\vx^1)$ relative to a fixed reference, not the transition
$g(\vx^1)-g(\vx^0)$ along a path. Equal-split counterfactual Shapley 
always splits interactions into equal contributions. SHAP explains a single prediction relative to a background distribution rather than a counterfactual transition, and provides no geometry awareness or interaction pot decomposition. 

Each missing property has a concrete cost: 
\textbf{CF} failure means the method targets point attribution, 
not transition attribution; \textbf{Geo} failure means 
within-pot splits default to equal shares, blind to which 
feature enables an interaction early along the path; 
\textbf{Int} failure means main effects and synergies are 
conflated; \textbf{IG~lim} failure means attributions do not 
converge to the continuous Aumann--Shapley integral under 
grid refinement; and \textbf{Poly} failure means computation 
scales exponentially in the number of changed features. 
\texttt{Aumann-SHAP} is the only method that satisfies all 
five simultaneously.


\section{TU-micro-game: decomposition of feature interactions}\label{sec:section3}

\subsection{Setup}\label{sec:setup}


Fix $d\ge 2$ and let $N:=\{1,\dots,d\}$ index the features. A coalition is any subset $S\subseteq N$.
Let $g:\mathcal{X}\to\mathbb{R}$ be a predictive functional on an input space $\mathcal{X}\subseteq\mathbb{R}^d$,
and let $\vx=(x_1,\dots,x_d)\in\mathcal{X}$ denote an instance.\footnote{If baseline/counterfactual objects are rows
of a data matrix $\vX\in\mathbb{R}^{n\times d}$, then $\vx^0,\vx^1$ can be read as two fixed rows (two instances),
and $\vx_u(t)$ is defined coordinatewise exactly as in \eqref{eq:slider_point}.}
Fix two endpoints $\vx^0,\vx^1\in\mathcal{X}$ (baseline and counterfactual). For any coalition $S\subseteq N$,
define the mixed input $\vx^{S}\in\mathcal{X}$ coordinatewise by
\begin{equation}\label{eq:mixed_input}
x^{S}_i :=
\begin{cases}
x^{1}_i, & i\in S,\\
x^{0}_i, & i\notin S.
\end{cases}
\end{equation}
Define the finite-change coalition value \cite{shapley1953,lundberg2017shap} by
$V(S):=g(\vx^{S})-g(\vx^{0})$ for $S\subseteq N$, so $V(\varnothing)=0$ and
\begin{equation}\label{eq:total_change}
\Delta y:=V(N)=g(\vx^{1})-g(\vx^{0}).
\end{equation}
For every nonempty $u\subseteq N$, define the M\"obius/Harsanyi dividend \cite{harsanyi1963} by
\begin{equation}\label{eq:mobius_dividend}
\phi_u := \sum_{T\subseteq u} (-1)^{|u|-|T|}\,V(T).
\end{equation}
All interaction pots in $S$ reconstruct $V$ (M\"obius inversion \cite{harsanyi1963}):
\begin{equation}\label{eq:mobius_reconstruction}
V(S)=\sum_{u\subseteq S}\phi_u \ \text{for all }S\subseteq N,
\qquad
\Delta y=\sum_{\varnothing\neq u\subseteq N}\phi_u .
\end{equation}

\begin{table}[H]
\centering
\scriptsize\caption{Notation used throughout the paper.}
\label{tab:notation}
\footnotesize
\setlength{\tabcolsep}{4pt}
\renewcommand{\arraystretch}{1.1}
\begin{tabular}{ll}
\toprule
\textbf{Symbol} & \textbf{Meaning}\\
\midrule
$N=\{1,\dots,d\}$ & full feature index set\\
$u\subseteq N$, $|u|=k$ & coalition of features under study\\
$T\subseteq u$ & sub-coalition (corner of local cube)\\
$\vx^0,\vx^1\in\mathcal{X}$ & baseline and counterfactual instance\\
$\vx^S$ & mixed input (\textit{cf.}~Eq.~\eqref{eq:mixed_input})\\
$V(S)=g(\vx^S)-g(\vx^0)$ & finite-change coalition value\\
$\phi_u$ & Harsanyi/M\"obius dividend of $u$\\
$\Delta y=g(\vx^1)-g(\vx^0)$ & total change being explained\\
$\mvec=(m_i)_{i\in u}$ & per-feature grid resolution\\
$r_u(t)$ & residual interaction surface\\
$p\in\mathcal{G}(u)$ & grid state\\
$N'_u$ & micro-player set, $|N'_u|=\sum_{i\in u}m_i$\\
$v_u(A)$ & TU-micro-game on $N'_u$\\
$S^{\mathrm{Sh}}_{i\to u}$, $S^{\mathrm{LES}}_{i\to u}$ & feature-level within-pot share\\
\bottomrule
\end{tabular}
\end{table}

\subsection{Counterfactual analysis on local cube restriction}\label{sec:local_cube}

Fix a nonempty coalition $u\subseteq N$ with $|u|=k$, let $\Delta_i:=x_i^{1}-x_i^{0}$, and let $(e_i)_{i\in N}$
be the canonical basis of $\mathbb{R}^d$. Let $t=(t_i)_{i\in u}\in[0,1]^k$, and define the cube-restricted path by
\begin{equation}\label{eq:slider_point}
\vx_u(t):=\vx^0+\sum_{i\in u} t_i\,\Delta_i\,e_i, \ t\in[0,1]^k,
\end{equation}
and set the model $g_u(t):=g(\vx_u(t))$. Only coordinates in $u$ vary continuously from baseline to counterfactual via $t_i\in[0,1]$.
Corners $t\in\{0,1\}^k$ correspond to full switches of selected features.
Evaluations on faces (masking) are obtained by fixing some coordinates at baseline, \textit{i.e.}, setting $t_i=0$ for
masked features.

\noindent\textit{Corners.}
For any $T\subseteq u$, define $t^{T}\in\{0,1\}^k$ by $(t^{T})_i=1$ if $i\in T$ and $(t^{T})_i=0$ otherwise.
Then $\vx_u(t^{T})=\vx^{T}$ and
$g_u\!\big(t^{T}\big)=g\!\big(\vx^{T}\big)$, for $T\subseteq u.$

\noindent\textit{Masking.}
For $T\subseteq u$ and $t\in[0,1]^k$, define the masked slider vector $t^{(T)}\in[0,1]^k$:
\[
(t^{(T)})_i :=
\begin{cases}
t_i, & i\in T,\\
0, & i\in u\setminus T.
\end{cases}
\]
Thus $g_u(t^{(T)})$ evaluates the face where only features in $T$ may move.

\begin{figure}[t]
\centering
\begin{subfigure}[t]{0.48\linewidth}
  \centering
  \includegraphics[width=\linewidth]{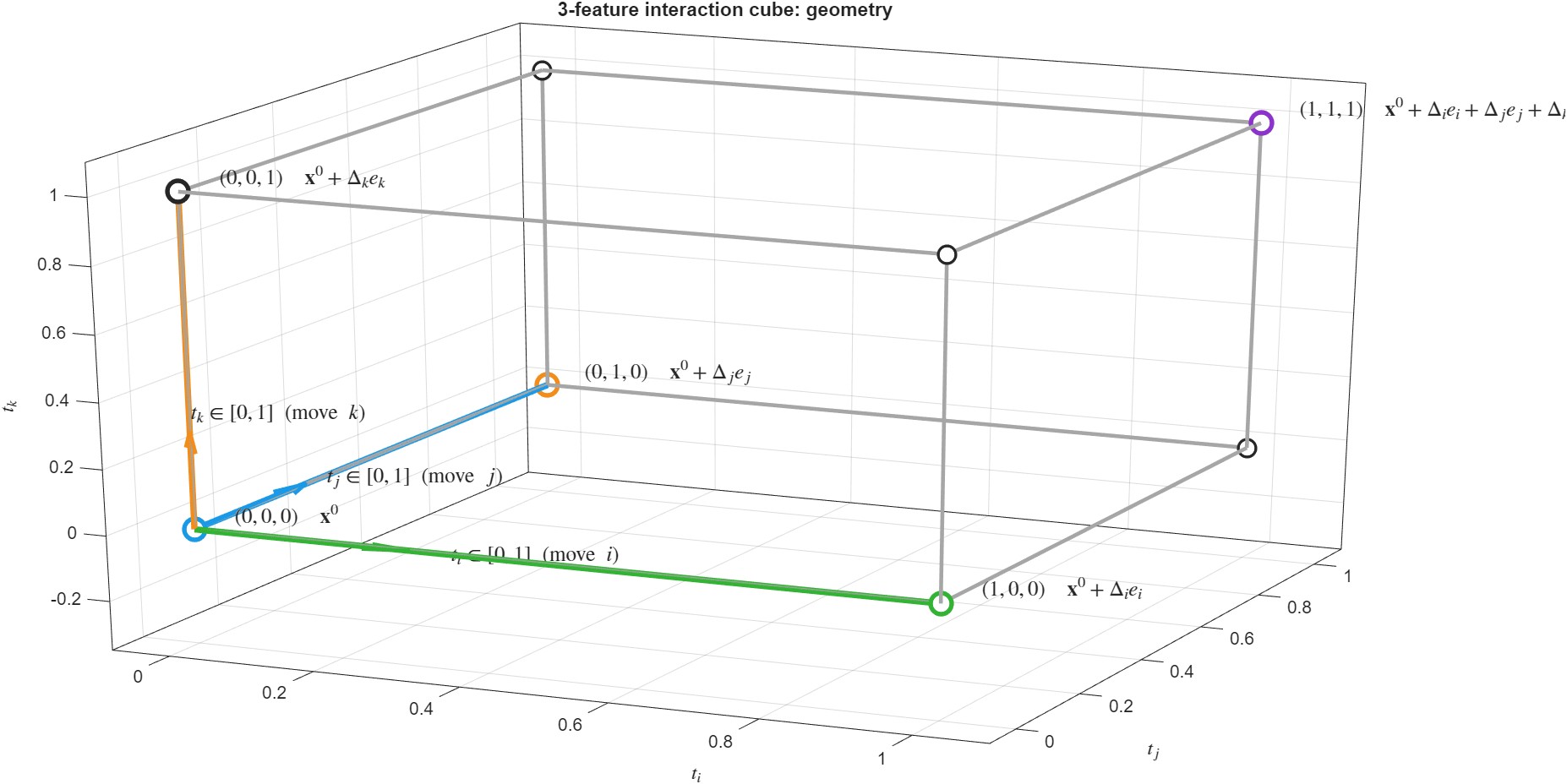}
  \caption{Cube corners.}
  \label{fig:Cube}
\end{subfigure}\hfill
\begin{subfigure}[t]{0.48\linewidth}
  \centering
  \includegraphics[width=\linewidth]{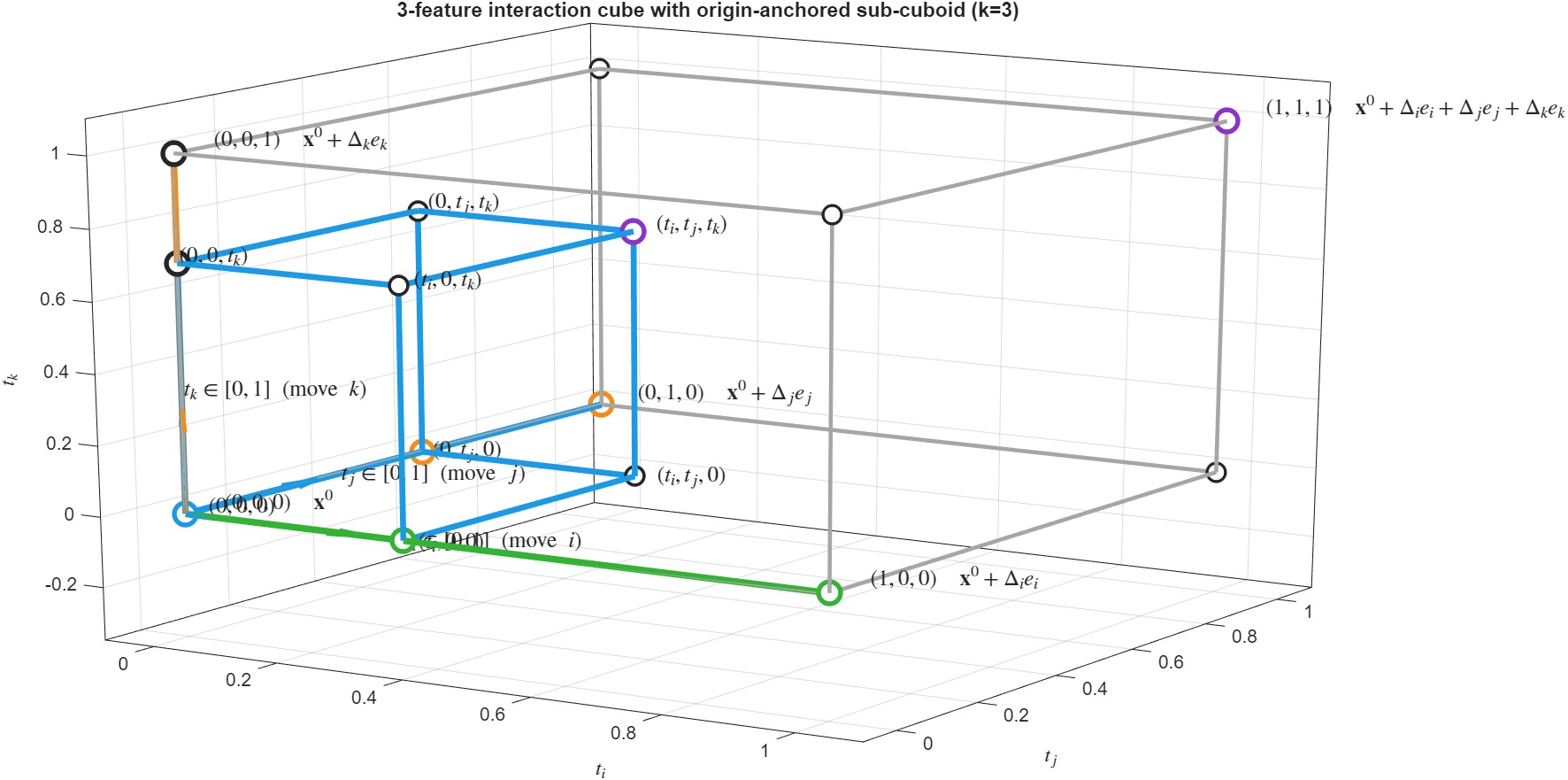}
  \caption{Sub-cube to $(t_i,t_j,t_k)$.}
  \label{fig:subcube}
\end{subfigure}
\caption{
\scriptsize\textbf{Local slider cube for $u=\{i,j,k\}$ (a) and the sub-cube from $(0,0,0)$ to $(t_i,t_j,t_k)$ (b), illustrating corner evaluations and the inclusion--exclusion construction.}}
\label{fig:cube_and_subcube}
\end{figure}

\begin{definition}[\textbf{Harsanyi dividend on a local-cube}]\label{def:pure_residual}
Fix $u\subseteq N$ with $|u|=k\ge 2$ and the cube-restricted model $g_u:[0,1]^k\to\mathbb{R}$.
The pure $u$-way residual interaction is the inclusion--exclusion surface
\begin{equation}\label{eq:residual_general}
r_u(t):=\sum_{T\subseteq u} (-1)^{k-|T|}\, g_u\!\big(t^{(T)}\big),\ t\in[0,1]^k.
\end{equation}
\end{definition}

\begin{proposition}[\textbf{Properties}]\label{prop:ru_boundary_corner}
Let $u\subseteq N$ with $|u|=k\ge 2$.
\newline $(\imath)$ If $t_i=0$ for some $i\in u$, then $r_u(t)=0$. \emph{\textcolor{blue}{(Boundary)}}
\newline $(\imath\imath)$ At the far corner, $r_u(\one_k)=\phi_u$. \emph{\textcolor{blue}{(Efficiency)}}
\end{proposition}

\noindent $(\imath)$ The joint presence of all $k$ features vanishes as soon as one is masked to baseline. $(\imath\imath)$ At the far corner the Harsanyi pot $\phi_u$ is recovered.

\subsection{Discretization of the cube on a grid \& TU-micro-game}\label{sec:grid_discretize}

The residual interaction $r_u(t)$ describes how the interaction between features builds up inside the cube.
To redistribute the interaction pot $\phi_u$ in a geometry-aware way, we discretize the cube and treat elementary step moves as players. Fix a resolution profile $\mvec=(m_i)_{i\in u}$ with $m_i\ge 1$ and restrict each slider to
$0,\frac1{m_i},\frac2{m_i},\dots,1$. A grid node is indexed by
\[
p=(p_i)_{i\in u}\in\prod_{i\in u}\{0,1,\dots,m_i\},\quad
t(p)=\Big(\frac{p_i}{m_i}\Big)_{i\in u}\in[0,1]^k.
\]

\begin{definition}[\textbf{Grid discretization}]\label{def:grid_residual}
	Fix $u\subseteq N$ with $|u|=k\ge 2$ and $\mvec=(m_i)_{i\in u}$ with $m_i\ge 1$.
	Define the sampled cube table $g_p := g_u(t(p))$. For $T\subseteq u$, define the masked index $p^{(T)}$ by keeping $p_i$ for $i\in T$
	and setting $p_i=0$ for $i\in u\setminus T$. The discrete residual interaction table is given by:
	\begin{equation}\label{eq:discrete_residual_rp}
		r_p := \sum_{T\subseteq u} (-1)^{k-|T|}\, g_{p^{(T)}}.
	\end{equation}
\end{definition}

\noindent The grid discretization is consistent, \textit{i.e.} the table $(r_p)$ is exactly the residual interaction surface sampled on the grid $r_p=r_u(t(p))$.
In particular, $r_p=0$ when any coordinate is still at baseline (boundary property), and $r_{\mvec}=\phi_u$ at the far corner.

Fix $i\in u$ and let $e_i$ be the unit vector in the $i$-direction in index space. For any $p$ with $p_i<m_i$, \textit{a local change} on the grid is given by:
\begin{equation}\notag
	\Delta_p(i):=r_{p+e_i}-r_p .
\end{equation}
This provides a local change in the pure interaction pot when feature $i$ advances by one discretization step,
given the current context $p$ (\textit{i.e.}, how far the other features have already moved).\footnote{When a coordinate in $u$ is categorical, intermediate values in \eqref{eq:slider_point} may not be valid inputs.
We therefore evaluate interior points by an endpoint-mixture that preserves all cube corners (and therefore preserves $V(\cdot)$ and the pots $(\phi_u)$).
For a single categorical coordinate $i\in u$, we define
$g_u^\star(t):=(1-t_i)\,g(\vx_{u,i}^{0}(t))+t_i\,g(\vx_{u,i}^{1}(t))$.
Thus $V(\cdot)$ and $(\phi_u)$ are unchanged, but the interior geometry is that of the mixture extension $g_u^\star$ rather than the raw model on invalid intermediate categorical inputs.}

\subsection{Micro-players and TU-micro-games on the grid}\label{sec:micro_players}
Discretization replaces each feature move $0\to 1$ by $m_i$ elementary steps. These steps are defined to be \textit{micro-players} on a TU-micro-game.

\begin{definition}[\textbf{Micro-players and TU-micro-game}]\label{def:micro_players}
Fix $u\subseteq N$ with $|u|=k\ge 2$ and $\mvec=(m_i)_{i\in u}$ with $m_i\ge 1$.
Define
$
N'_u := \bigcup_{i\in u}\big(\{i\}\times\{1,\dots,m_i\}\big),
$
so $|N'_u|=\sum_{i\in u}m_i$.
For a micro-coalition $A\subseteq N'_u$, define $c_i(A):=\big|\{\,s\in\{1,\dots,m_i\}:\ (i,s)\in A\,\}\big|\in\{0,1,\dots,m_i\}$.
The TU-micro-game is defined as:
$$
v_u:\;2^{N'_u}\to\mathbb{R}, \ \text{ such that } v_u(A):=r_{p(A)} \ \text{ and } v_u(\varnothing)=0.
$$
\end{definition}

\noindent For all moves, $v_u(N'_u)=r_{\mvec}=\phi_u$, the interaction pot is recovered. For $(i,s)\notin A$, adding the micro-player increases $c_i(A)$ by one, hence moves the state from $p(A)$ to
$p(A)+e_i$ (with $p(A):=\big(c_i(A)\big)_{i\in u}$). Therefore the marginal contribution of a micro-player equals the residual interaction increment:
\[
v_u\big(A\cup\{(i,s)\}\big)-v_u(A)=r_{p(A)+e_i}-r_{p(A)}=\Delta_{p(A)}(i).
\]
This identity links within-cube interaction geometry (the increments $\Delta_p(i)$) to cooperative-game allocation on
$v_u$, enabling feature-level attributions by aggregating micro-player payoffs (Section~\ref{sec:Aumann-SHAP}).

\section{Aumann-SHAP}\label{sec:Aumann-SHAP}

\subsection{Interaction, local and global explainability}

A value (attribution method) applied to $v_u$ produces payoffs for micro-game players (step moves).
A \emph{feature-level} within-pot attribution is built by summing the micro-players that belong to the same feature. A value is a function $\varphi$ on $\Gamma^u$, \textit{i.e.} $\varphi:\Gamma^u\to\mathbb{R}^{N'_u}$,
that assigns a payoff vector $\varphi(v)=(\varphi_a(v))_{a\in N'_u}$ to each TU-micro-game $v_u\in\Gamma^u$.
The Shapley value \cite{shapley1953} is the unique value satisfying Efficiency (\textbf{E}), Symmetry (\textbf{S}), Additivity (\textbf{A}), and the Null player (\textbf{N}).\footnote{\textbf{Efficiency (E).} For each $v_u\in\Gamma^u$, $\sum_{a\in N'_u}\varphi_a(v_u)=v_u(N'_u)$.
\textbf{Symmetry (S).} For each $v_u\in\Gamma^u$, for any permutation $\sigma$ of $N'_u$, define $(\sigma v_u)(A):=v_u(\sigma^{-1}(A))$. Then, for all $a\in N'_u$, $\varphi_{\sigma(a)}(\sigma v_u)=\varphi_a(v_u)$.
\textbf{Additivity (A).} For each $v_u,v'_u\in\Gamma^u$, $\varphi(v_u+v'_u)=\varphi(v_u)+\varphi(v'_u)$.
\textbf{Null player (N).} For each $v_u\in\Gamma^u$, if $a$ is null, i.e. $v_u(S\cup\{a\})=v_u(S)$ for all $S\subseteq N'_u\setminus\{a\}$, then $\varphi_a(v_u)=0$.}

\begin{theorem}[\textbf{Micro-player Shapley value}]\label{thm:micro_shapley}
A value $\varphi$ on $\Gamma^u$ satisfies \textbf{E}, \textbf{S}, \textbf{A} and \textbf{N} if, and only if, for every micro-player $(i,s)\in N'_u$,
\begin{multline*}
\varphi^{\mathrm{Sh}}_{(i,s)}(v_u)
=
\sum_{A\subseteq N'_u\setminus\{(i,s)\}}
\frac{|A|!\,(|N'_u|-|A|-1)!}{|N'_u|!}\\
\times\Big(
v_u(A\cup\{(i,s)\})-v_u(A)
\Big).
\end{multline*}
\end{theorem}

\textit{Interaction explainability.}
The term inside parenthesis is the marginal increase in interaction when one elementary step is added. The marginal moves are driven by the local increments $\Delta_p(i)$, therefore the Shapley value $\varphi^{\mathrm{Sh}}_{(i,s)}$ is the average of all feature interactions over the sub-cube (see Appendix for the proofs of theorems and propositions). For each feature $i\in u$, we define the feature-level interaction contribution by aggregation:
\[
S^{\mathrm{Sh}}_{i\to u}
:=
\sum_{s=1}^{m_i}\varphi^{\mathrm{Sh}}_{(i,s)}(v_u).
\]
Then, $S^{\mathrm{Sh}}_{i\to u}$ yields the contribution of feature $i$ to the entire interaction pot $\phi_u$.

\textit{Local and global explainability.}
Local explainability is issued from all possible within-pot shares including feature $i$, \textit{i.e.} $S^{\mathrm{Sh}}_{i\to u}$ (for all $u$), and the contribution of $i$ alone, for one given instance:
\[
S_i^{\mathrm{Sh}}
\;:=\;
\phi_{\{i\}}
\;+\;
\sum_{\substack{u\subseteq N:\; i\in u\\ |u|\ge 2}}
S^{\mathrm{Sh}}_{i\to u}.
\]
This yields the explained variation $g(\vx^{1})-g(\vx^{0})$ for one given instance of the test data. Global explainability (for each feature $i$) is obtained by averaging local contributions, over the test data, in which $i$ participates. By efficiency, the global feature explainabilities sum to the explained variation on test data.

\begin{proposition}[\textbf{Equal-split Shapley }\cite{luborgonovo2024}]\label{prop:m1_equal_split}
	If $m_i=1$ for every $i\in u$, then $v_u(A)=0$ for all $A\subsetneq N'_u$ and $v_u(N'_u)=\phi_u$.
	Therefore,
	\[
	S_i^{\mathrm{Sh}}(u)=\frac{\phi_u}{|u|}, \ \forall i\in u.
	\]
\end{proposition}

\noindent Proposition~\ref{prop:m1_equal_split} shows that the result of \cite{luborgonovo2024} is a special case for one single move $m_i=1$. This particular case is less informative: each feature $i$ in the interaction pot $\phi_u$ receives exactly the same contribution. Refining the grid on a continuum (large $m$) avoids this problem and produces an Aumann-Shapley path-based attribution method on the residual interaction surface.

\begin{theorem}[\textbf{Aumann--SHAP}]\label{thm:m_to_infty_diagonal_AS}
	Fix $u\subseteq N$ with $|u|=k\ge2$ and a resolution profile $\mvec=(m_i)_{i\in u}$.
	Assume $r_u\in C^1([0,1]^k)$. Then, for every $i\in u$,
	\begin{equation}\notag
		\lim_{\min_{j\in u} m_j\to\infty}\mathrm{S}^{\mathrm{Sh}}_{i \to u}
		=
		\int_{0}^{1}\frac{\partial r_u}{\partial t_i}\big(\tau\,\one_k\big)\,d\tau.
	\end{equation}
	Moreover, these limit contributions are efficient, $\sum_{i\in u}\int_{0}^{1}\frac{\partial r_u}{\partial t_i}\big(\tau\,\one_k\big)\,d\tau = \phi_u$.\footnote{Theorem~\ref{thm:m_to_infty_diagonal_AS} should be read as a smooth-case limit result. For non-smooth models (e.g. tree ensembles), our attribution method is still defined through finite grid evaluations of $r_u$, while convergence to a path-integral limit is left outside the scope of the theorem.}
\end{theorem}

\subsection{Ground-truth validation}

\begin{figure}[h!]
\centering
\includegraphics[width=\linewidth]{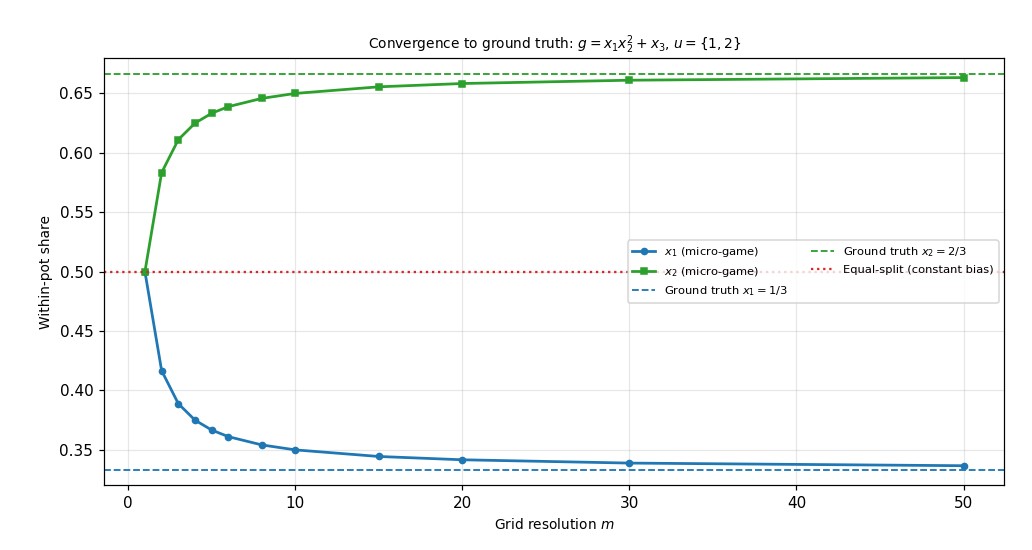}
\caption{\scriptsize\textbf{Convergence to analytical ground truth.
Micro-game Shapley (solid) converges to the exact $1/3$--$2/3$
split (dashed) as $m$ increases. Equal-split Shapley (dotted red)
remains at $0.5$ for all $m$ — a permanent, irreducible bias.
Model: $g=x_1x_2^2+x_3$, $u=\{1,2\}$,
$\mathbf x^0=(0,0,0)\to\mathbf x^1=(1,1,1)$.}}
\label{fig:convergence}
\end{figure}

On the synthetic model $g(\mathbf x)=x_1x_2^2+x_3$ with transition
$\mathbf x^0=(0,0,0)\to\mathbf x^1=(1,1,1)$, the interaction pot
for $u=\{1,2\}$ is $\phi_u=1$ and the diagonal Aumann--Shapley
integral yields analytically exact shares $1/3$ for $x_1$ and
$2/3$ for $x_2$: feature~$2$'s quadratic term drives interaction
growth faster along the path. Equal-split Shapley assigns $0.5$
each
, as it is blind to
the asymmetric curvature. Fig.~\ref{fig:convergence} shows that
our grid-state Shapley converges monotonically to the $1/3$--$2/3$
ground truth as $m$ increases (reaching $0.333$/$0.667$ at $m=50$),
while equal-split remains at $0.5$ regardless of micro-game, \textit{i.e. micro-variation of the features}.
This is not a numerical artifact: the convergence is guaranteed
by Theorem~\ref{thm:m_to_infty_diagonal_AS}, and the equal-split
bias is structural, it cannot be removed by any grid refinement.

\subsection{Generalization to LES values}

Micro-game-Shapley is one particular way to split a fixed interaction pot $\phi_u$ across its member features. Linear--Efficient--Symmetric (LES) values enable more attribution methods to be employed. Taking recourse to LES (in particular Linearity (\textbf{L})\footnote{\textbf{Linearity (L).}
For each $v_u,w_u\in\Gamma^u$ and $\alpha_1,\alpha_2\in\mathbb{R}$,
$\varphi(\alpha_1v_u+\alpha_2w_u)=\alpha_1\varphi(v_u)+\alpha_2\varphi(w_u)$.} instead of Additivity) enables the explained quantity fixed (pots remain the same), while \emph{within-pot split} are differently evaluated (for robustness purpose).

\noindent
Invoking \textbf{L}, \textbf{E}, and \textbf{S} implies (and is implied) by the LES family:
\begin{multline*}
\varphi^{\mathrm{LES}}_{(i,s)}(v_u)
=
\sum_{A\subseteq N'_u\setminus\{(i,s)\}}
\frac{|A|!\,(n-|A|-1)!}{n!}\\
\times\Big(b(|A|+1)\,v_u(A\cup\{(i,s)\})-b(|A|)\,v_u(A)\Big),
\end{multline*}
where $n:=|N'_u|$ and $b(\cdot)$ is a scalar sequence with $b(0)=0$ and $b(n)=1$, see
\cite{ruizvalencianozarzuelo1998,chameninembuaandjiga2008,chameninembua2012,radzikdriessen2013,bealremilasolal2015}.
Aggregating micro-players back to a feature-level yields:
\[
S^{\mathrm{LES}}_{i\to u}:=\sum_{s=1}^{m_i}\varphi^{\mathrm{LES}}_{(i,s)}(v_u),\ \forall i\in u.
\]
By efficiency, within-pot share gives \textit{interaction explainability} $S^{\mathrm{LES}}_{i\to u}$. \textit{Local explainability} is given by
$\phi_{\{i\}}+\sum_{\substack{u\subseteq N:\; i\in u\\ |u|\ge 2}} S^{\mathrm{LES}}_{i\to u}=S^{\mathrm{LES}}_{i}$ for one given instance (whereas \textit{global explainability} is obtained by averaging on test data). LES attribution methods change only \emph{how} $\phi_u$ is divided, not what is being explained. As shown by \cite{Condevaux2022}, comparing several LES attribution methods reports explanations that are stable across rules
(\textit{e.g.}, agreement on top features), and may provide faster computation rule to be employed such as ES (see Table~\ref{tab:les_three_complexities}).

\begin{table}[t]
\centering
\caption{\scriptsize\textbf{Three LES instantiations ($n=\sum_{j\in u}m_j$, $k=|u|$).}}
\label{tab:les_three_complexities}
\small
\setlength{\tabcolsep}{5pt}
\renewcommand{\arraystretch}{1.1}
\begin{adjustbox}{max width=\linewidth}
  \begin{tabular}{l c c}
    \toprule
    LES value & Choice of $b(s)$ (for $1\le s\le n-1$) & Complexity (per pot $u$)\\
    \midrule
    Shapley value & $b(s)=1$ & $\mathcal{O}\!\big(k(m+1)^k\big)$\\
    Solidarity value & $b(s)=\frac{1}{s+1}$ & $\mathcal{O}\!\big(k(m+1)^k\big)$\\
    Equal Surplus (ES) & $b(1)=n-1,\; b(s)=0\ (s\ge2)$ & $\mathcal{O}(k)$\\
    \bottomrule
  \end{tabular}
\end{adjustbox}
\end{table}

\section{Experiments}\label{sec:experiments}

We evaluate \texttt{Aumann-SHAP} on four benchmarks: complexity simulations, German Credit \cite{ustun2019actionable}, UCI Heart Disease \cite{uci_heart}, and MNIST \cite{lecun2010mnist}.\footnote{Code: \url{https://github.com/icdm-anon-2026/Aumann-SHAP}.}

\subsection{Simulations: Exact computation and complexity}

Direct computation of Shapley/LES on $v_u$ over $N'_u$ (with $n=\sum_{j\in u}m_j$ micro-players) requires summing
over $2^n$ micro-coalitions, \textit{i.e.} $\mathcal{O}(n\,2^n)$ operations per interaction pot. The key structure of our induced TU-micro-game is $v_u(A)=r_{p(A)}$,
therefore the game $v_u$ depends only on its \emph{grid state} $p(A)$. We can therefore group terms by states by defining $\Delta_p(i):=b(|p|+1)r_{p+e_i}-b(|p|)r_p$, if $p_i<m_i$, and $0$ otherwise.

\begin{proposition}[\textbf{LES closed form}]\label{prop:grid_state_closed_form}
Fix $u\subseteq N$ with $|u|\ge 2$ and $\mvec=(m_j)_{j\in u}$, and let $n=\sum_{j\in u}m_j$. Then, for every $i\in u$,
\begin{multline}\label{eq:share_grid_les_closed_form}
S^{\mathrm{LES}}_{i\to u}
=
\sum_{\substack{p\in\mathcal{G}(u)\\ p_i<m_i}}
\frac{|p|!\,(n-|p|-1)!}{n!}\\
\times\Bigg(\prod_{j\in u}\binom{m_j}{p_j}\Bigg)\,
(m_i-p_i)\,\Delta_p(i).
\end{multline}
\end{proposition}

\begin{figure}[H]
\centering
\begin{subfigure}[t]{0.48\linewidth}
\centering
\includegraphics[width=\linewidth]{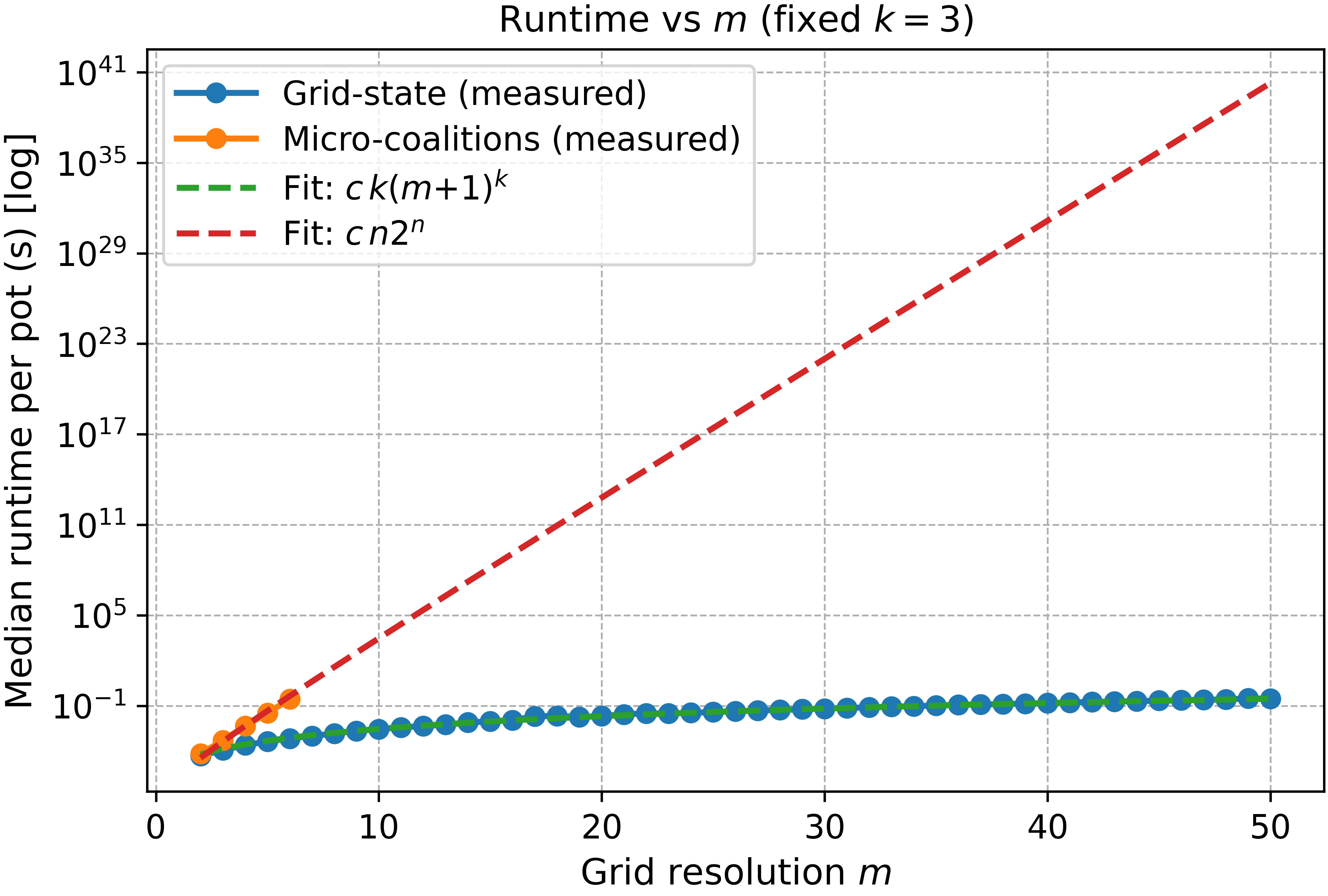}
\caption{Runtime vs.\ $m$.}
\label{fig:runtime_vs_m}
\end{subfigure}\hfill
\begin{subfigure}[t]{0.48\linewidth}
\centering
\includegraphics[width=\linewidth]{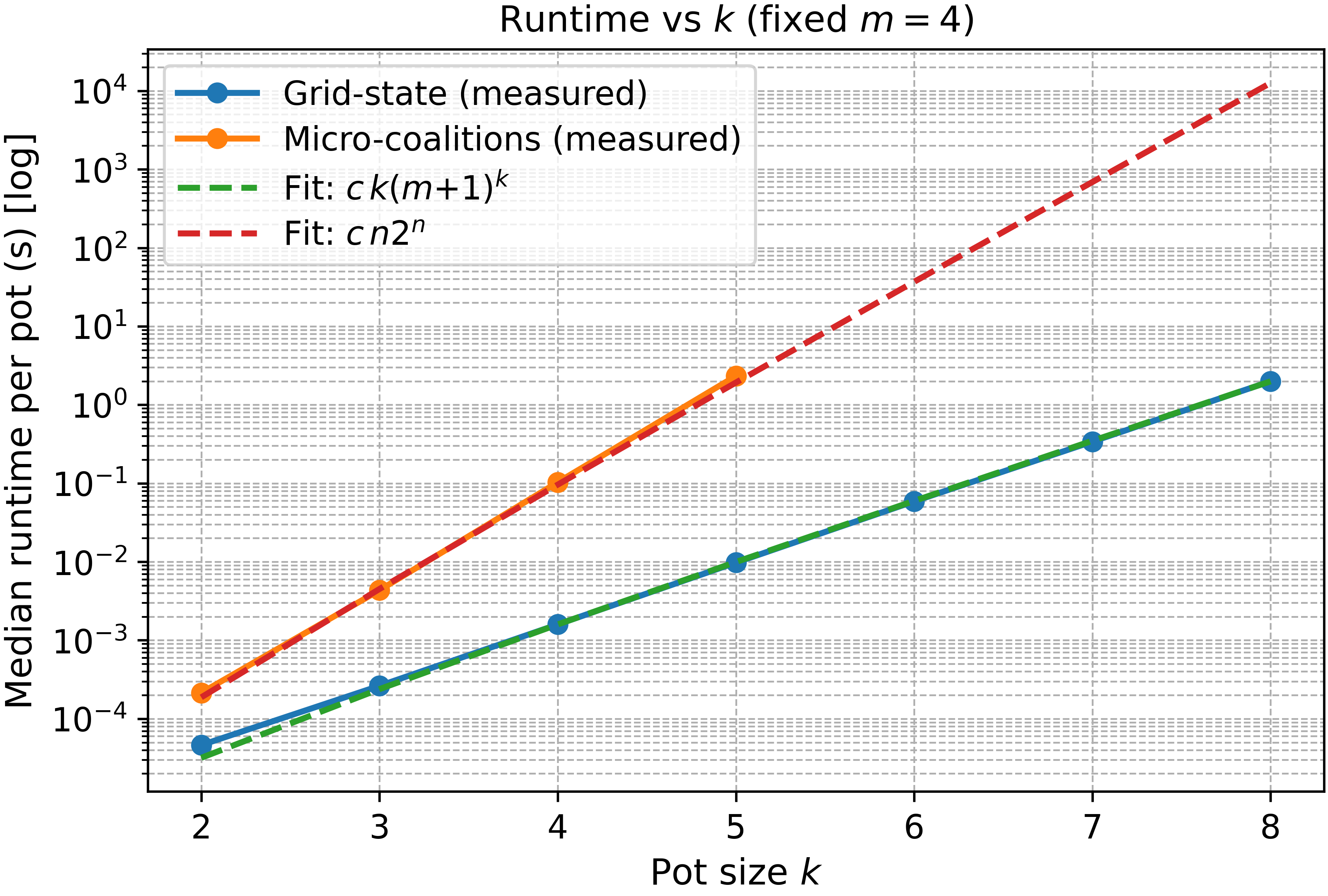}
\caption{Runtime vs.\ $k$.}
\label{fig:runtime_vs_k}
\end{subfigure}
\caption{
\scriptsize
\textbf{Solid curves with markers provide runtimes: grid-state computation (blue) and, when feasible, full micro-coalition enumeration (orange). Dashed curves are fitted scaling laws: $\sim c\,k(m{+}1)^k$ for grid-state (green) and $\sim c\,n2^n$ with $n{=}km$ for full enumeration (red).}}
\label{fig:runtime_two_panels}
\end{figure}

\textit{Complexity.} In the uniform-resolution regime $m_j\equiv m$ (so $n=km$), full enumeration costs
$\mathcal{O}(km\,2^{km})$.
Eq.~\eqref{eq:share_grid_les_closed_form} loops over $(m+1)^k$ states and $k$ directions, yielding $\mathcal{O}\!\big(k(m+1)^k\big)$,
polynomial in $m$ for fixed $k$ (and exponential only in $k$). Equal Surplus is $\mathcal{O}(k)$ once the needed
corner quantities are computed \cite{Condevaux2022}. Empirically, the reduction is decisive as shown in Fig.~\ref{fig:runtime_two_panels}. At fixed $k=3$, increasing $m$
quickly makes full enumeration infeasible, while the grid-state method remains fast. At fixed $m=4$, grid-state
runtimes grow as $(m+1)^k$ while full enumeration grows as $2^{km}$.

\begin{algorithm}[t]
\caption{Grid-state LES share $S^{\mathrm{LES}}_{i\to u}$}
\label{alg:grid_les}
\begin{algorithmic}[1]
\Require model $g$, endpoints $\vx^0,\vx^1$, coalition $u$, resolution $\mvec$, weights $b(\cdot)$
\Ensure feature-level shares $\{S^{\mathrm{LES}}_{i\to u}\}_{i\in u}$
\State Tabulate $g_p\gets g_u(t(p))$ for all $p\in\mathcal{G}(u)$
\State Compute $r_p\gets\sum_{T\subseteq u}(-1)^{k-|T|}g_{p^{(T)}}$ \Comment{Eq.~\eqref{eq:discrete_residual_rp}}
\ForAll{$i\in u$}
  \State $S^{\mathrm{LES}}_{i\to u}\gets 0$
  \ForAll{$p\in\mathcal{G}(u)$ with $p_i<m_i$}
    \State $\Delta\gets b(|p|+1)r_{p+e_i}-b(|p|)r_p$
    \State $w\gets\tfrac{|p|!(n-|p|-1)!}{n!}\prod_{j\in u}\binom{m_j}{p_j}(m_i-p_i)$
    \State $S^{\mathrm{LES}}_{i\to u}\gets S^{\mathrm{LES}}_{i\to u}+w\,\Delta$
  \EndFor
\EndFor
\end{algorithmic}
\end{algorithm}

\emph{Convergence.} The Riemann sum converges to the Aumann--Shapley integral at rate $\mathcal O(1/n)$, which can also depend on model regularity along the interpolation path. For smooth models (\textit{e.g.} GLM see Table~\ref{tab:m_reco_models}), residual interaction increments vary gradually across states and moderate $m$ is sufficient. For
threshold predictors (tree ensembles, boosted trees), attributions may change when refinement resolves new split
thresholds, larger $m$ improves stability while remaining tractable \textit{via} the grid-state formula.\footnote{We apply a conservative saturation rule: increase $m$ until the maximum absolute change in any feature share
between successive $m$ values is below a tolerance $\varepsilon$ (\textit{e.g}., $0.1$ percentage points) for a few
consecutive refinements (\textit{e.g}., three), and report the resulting attributions.}

\begin{table}[t]
\centering
\caption{\scriptsize\textbf{Typical grid resolutions required for stable attributions in our experiments.}}
\label{tab:m_reco_models}
\footnotesize
\setlength{\tabcolsep}{5pt}
\renewcommand{\arraystretch}{1}
\begin{adjustbox}{max width=\linewidth}
\begin{tabular}{l c c}
\toprule
Model class & Typical $m$ & Default recommendation \\
\midrule
GLM regressions (smooth, linear) & $m \approx 4$--$6$ & start at $m=5$ \\
MLP (smooth, differentiable) & $m \approx 6$--$10$ & start at $m=8$ \\
XGBoost / GBDT (thresholded, piecewise) & $m \approx 8$--$15$ & start at $m=10$ \\
\bottomrule
\end{tabular}
\end{adjustbox}
\end{table}

\subsection{UCI Heart Disease}

On the UCI Cleveland Heart Disease dataset~\cite{uci_heart} ($n=303$, $13$ features, XGBoost AUC$=0.91$), we select a counterfactual pair with $k=3$ changed numeric features on the interaction pot $u=\{\texttt{chol},\texttt{thalach},\texttt{oldpeak}\}$\footnote{%
\texttt{chol}: serum cholesterol (mg/dl);
\texttt{thalach}: maximum heart rate achieved during exercise stress test (bpm);
\texttt{oldpeak}: ST depression induced by exercise relative to rest (mm).%
}, yielding $\phi_u=0.0103$. The baseline patient (no disease, label 0) has $\texttt{chol}=240$, $\texttt{thalach}=171$, $\texttt{oldpeak}=0.9$; the counterfactual patient (disease, label 1) has $\texttt{chol}=258$, $\texttt{thalach}=157$, $\texttt{oldpeak}=2.6$. The path is therefore from $(240,171,0.9)$ toward $(258,157,2.6)$: no disease $\rightarrow$ disease.

\begin{figure}[H]
\centering
\includegraphics[width=0.75\linewidth]{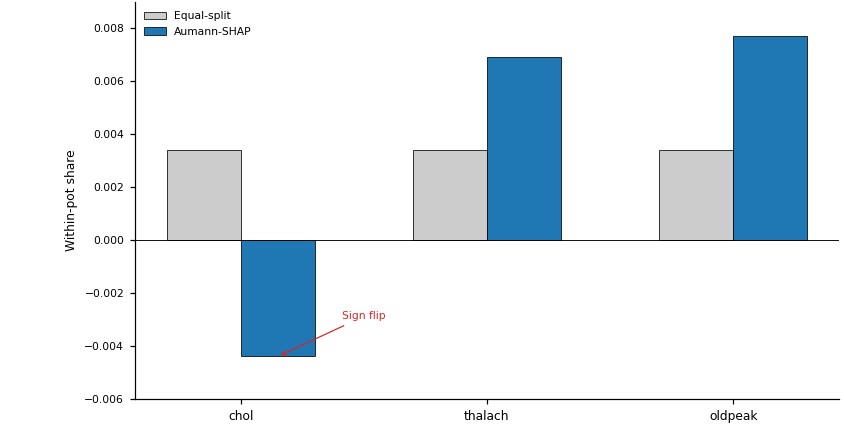}
\caption{
\scriptsize
\textbf{Medical misattribution risk. Equal-split 
(gray) assigns positive credit to \texttt{chol}, implying it 
contributes to the beneficial interaction. Aumann-SHAP (blue) 
reveals \texttt{chol} actively suppresses the interaction 
(negative share). A geometry-agnostic explanation could 
misdirect clinical intervention toward the wrong feature.}}
\label{fig:heart}
\end{figure}

The consequences of geometry-agnostic attribution in a medical 
setting are severe. Figure~\ref{fig:heart} illustrates the medical misattribution risk. Equal-split Shapley reports \texttt{chol} as a positive contributor ($+0.0034$) to the interaction, implying that increases in cholesterol reinforce the prediction transition toward disease.
 Aumann-SHAP recovers the opposite sign ($-0.0044$): 
\texttt{chol} actively suppresses the interaction, while 
\texttt{thalach} ($+0.0069$) and \texttt{oldpeak} ($+0.0077$) 
harvest the synergy. Notably, cholesterol \emph{increases} from 
240 to 258 along the path toward disease, yet its negative 
within-pot share means that \emph{reducing} cholesterol back 
toward the baseline suppresses the disease-driving interaction. 
An explanation used for clinical recourse or regulatory audit 
that misattributes a suppressor as a driver does not merely 
reduce fidelity, it risks misdirecting intervention toward the 
wrong physiological lever.

\subsection{German credit}\label{sec:german_credit}

The German Credit dataset \cite{hofmann1994statlog} contains 27 attributes for 1,000 applicants, with labels \emph{creditworthy} vs.\ \emph{not creditworthy} \cite{ustun2019actionable,luborgonovo2024}. We explain counterfactual transitions from a baseline $\vx^0$ (not creditworthy) to a successful counterfactual $\vx^1$ (creditworthy), outlining interaction responsibility among changed features.

Three classifiers (logistic regression, MLP, XGBoost) are combined into a conservative score $p_{\min 3}(\mathbf x)=\min\{p_{\text{logistic}},p_{\text{mlp}},p_{\text{xgboost}}\}$. We focus on a representative individual (idx 242, $p_{\min 3}=0.144$) and generate three successful counterfactuals via DiCE-style search, Growing Spheres, and a genetic algorithm, all with strict success condition $p_{\min 3}(\mathbf x^1)\ge 0.8$. Differences across explanations are \emph{not} about success vs.\ failure, but about how the gain $\Delta V$ is attributed.

Across all three counterfactuals, the top contributor is stable: $X_8$ dominates in Growing Spheres and Genetic, $X_{22}$ in DiCE. The method disagreement is never about \emph{which} feature matters most, but \emph{how much} the interaction geometry reallocates among secondary features.

\emph{Local explainability} (\cref{tab:totals_eq_vs_micro_dice}). We explain $\Delta V=0.6569$ (Growing Spheres, $u=\{X_8,X_{13},X_{14},X_{25}\}$) via equal-split Shapley $\phi^{\mathrm{eq}}$, micro-game Shapley $S^{\mathrm{micro}}$ ($m=5$), and Equal Surplus (ES). The dominant contributor is $X_8$. Beyond the top feature, the ranking of $X_{13}$ and $X_{14}$ differs: equal-split assigns more to $X_{13}$, whereas $S^{\mathrm{micro}}$ and ES assign more to $X_{14}$. The gap is driven by interaction geometry: equal-split enforces symmetry, while micro-game Shapley redistributes according to how gains accumulate inside the local cube.\footnote{ES is used only as a feature-total baseline, not as a canonical within-pot decomposition.}

\begin{table}[h!]
\centering
\caption{
\scriptsize
\textbf{Feature-level totals under equal splitting ($\phi^{\mathrm{eq}}$), TU-micro-game Shapley ($S^{\mathrm{micro}}$), and Equal Surplus ($ES$). Columns 4--5: \% computed on $\Delta V$. Growing Spheres CF, $\Delta V=0.6569$.}}
\label{tab:totals_eq_vs_micro_dice}
\footnotesize
\setlength{\tabcolsep}{4pt}
\renewcommand{\arraystretch}{1.15}
\begin{adjustbox}{max width=\linewidth}
\begin{tabular}{l c c c c c}
\toprule
\textbf{Feature} & $\boldsymbol{\phi^{\mathrm{eq}}}$ & $\boldsymbol{S^{\mathrm{micro}}}$ & $\boldsymbol{ES}$ &
$\boldsymbol{S^{\mathrm{micro}}-\phi^{\mathrm{eq}}}$ & $\boldsymbol{S^{\mathrm{micro}}-ES}$\\
\midrule
$X_8$ (\emph{Checking})        & 0.3331 & 0.4389 & 0.3013 & +\textcolor{blue}{16.11\%} & +20.95\%\\
$X_{14}$ (\emph{Loans Elsew.}) & 0.1295 & 0.1130 & 0.1576 & -2.50\%  & -6.79\%\\
$X_{13}$ (\emph{Credit hist.}) & 0.1772 & 0.0934 & 0.1313 & \textcolor{red}{-12.77\%} & -5.77\%\\
$X_{25}$ (\emph{Housing})      & 0.0171 & 0.0115 & 0.0666 & -0.85\%  & -8.39\%\\
\bottomrule
\end{tabular}
\end{adjustbox}
\end{table}

\emph{Interaction explainability} (\cref{tab:within_top_realloc}). Micro-game Shapley shifts substantial gain toward $X_{8}$ and away from $X_{13}$. In the $X_8$+$X_{13}$ pot, $X_{13}$ (\emph{Credit history}) negatively contributes (0.0176) while $X_{8}$ (\emph{Checking account}) drives improvement (0.1143). This single reallocation equals $7.4\%$ of the total gain $\Delta V$—a geometry-agnostic method would miss it entirely. Equal-split Shapley is silent on such interaction effects.

\begin{table}[H]
\centering
\caption{\scriptsize\textbf{Growing Spheres: within-pot reallocations (TU-micro-game vs.\ equal split).}}
\label{tab:within_top_realloc}
\small
\setlength{\tabcolsep}{5pt}
\renewcommand{\arraystretch}{1.15}
\begin{tabular}{l c c c c}
\toprule
\textbf{Pot $T$} & \textbf{Feat.} & \textbf{$\phi^{\mathrm{eq}}$} & \textbf{$S^{\mathrm{micro}}$} &
 \textbf{$\phi^{\mathrm{eq}}-S^{\mathrm{micro}}$} \\
\midrule
$X_{8}+X_{13}$ & $X_{13}$ & 0.0659 & \textcolor{blue}{0.0176} & \textbf{-0.0484}\\
$X_{8}+X_{13}$ & $X_{8}$  & 0.0659 & \textcolor{red}{0.1143} & \textbf{0.0484}\\
$X_{8}+X_{14}$ & $X_{14}$ & -0.0004 & -0.0366 & \textbf{-0.0362}\\
$X_{8}+X_{14}$ & $X_{8}$  & -0.0004 & 0.0359 & \textbf{0.0362}\\
\bottomrule
\end{tabular}
\end{table}

\emph{Global explainability.}
At the dataset level we aggregate over $n=138$ low-score baselines
($p_{\text{xgboost}}<0.30$) with successful counterfactuals
($p_{\text{xgboost}}(\vx^1)\ge 0.80$, $\bar{k}=2.51$ changed
features). \Xtwo\ and \Xone\ dominate average improvement under
all three rules. As a faithfulness metric, we measure the
\emph{rank-flip rate}: the fraction of instances where micro-game
Shapley produces a strictly different feature priority ordering
than equal-split Shapley. The rate is $12.3\%$ ($17/138$
instances), concentrated in cases with large interaction pots
where geometry awareness materially changes which lever a client
should act on first. In the remaining $87.7\%$ of instances the
two methods agree on feature priority, confirming robustness:
Aumann-SHAP refines the equal-split baseline where interactions
are strong, and recovers it otherwise.

\subsection{MNIST}

MNIST data \cite{lecun2010mnist} are used to visualize which pixels (and which pixel interactions) drive a counterfactual transition such as $1 \to 7$ (70,000 grayscale digits split into 60,000 train and 10,000 test samples). A ResNet-18 \cite{he2016resnet} architecture is trained and the resulting model attains a test accuracy of $0.9972$. To construct a counterfactual transition, we select a baseline image $\mathbf x^0$ of class $1$ and a target image $\mathbf x^1$ of class $7$, and we restrict the feature set $u$ to 109 pixels that actually differ between the two endpoints, \textit{i.e.} those satisfying $|x^1_{row,col}-x^0_{row,col}|>0.05$.

\begin{figure}[t]
\centering
\includegraphics[width=\linewidth]{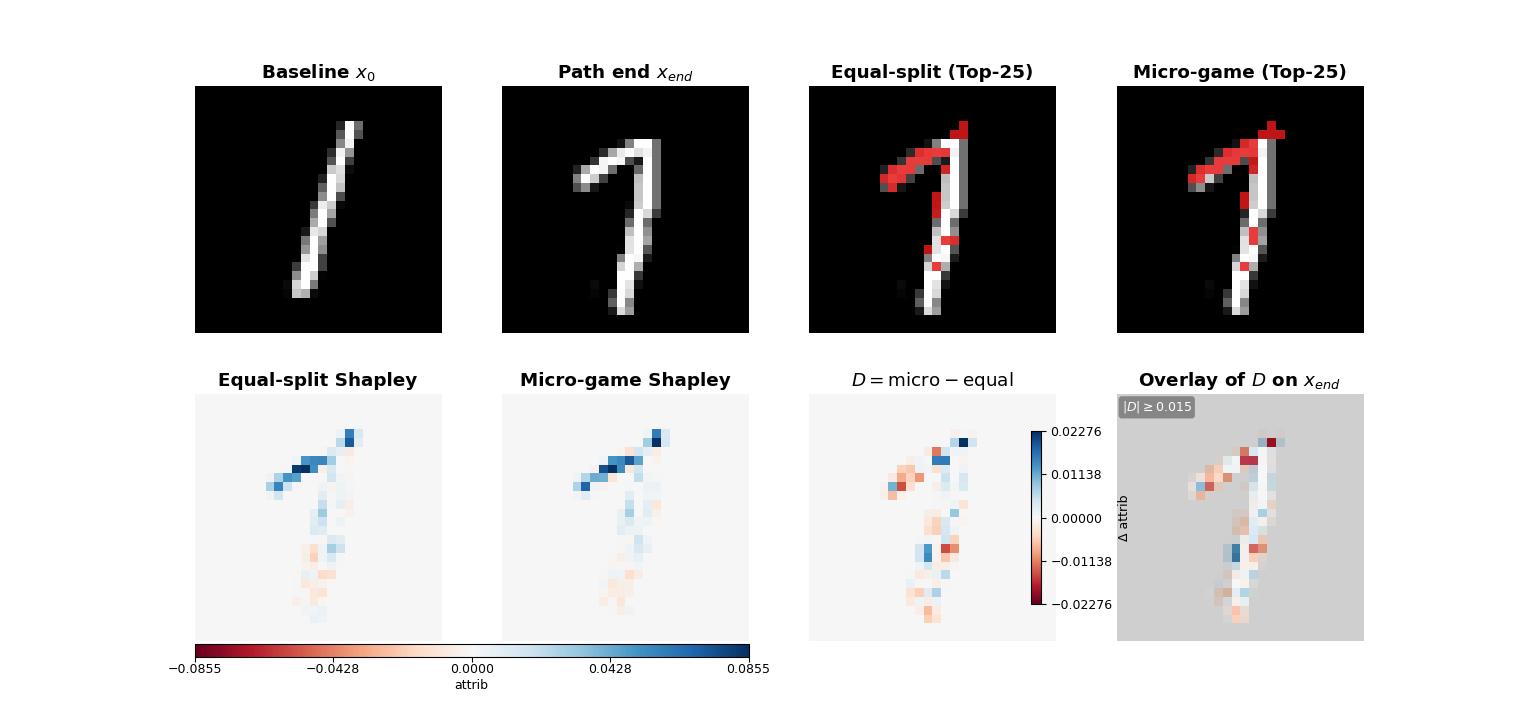}
\caption{
\scriptsize
\textbf{MNIST $1\to 7$ counterfactual: attribution maps and redistribution.
Top row: baseline $x^0$, endpoint $x^1$ (only the $k$ changed pixels are moved), and Top-25 masks for equal-split and TU-micro-game Shapley (red).
Bottom row: signed heatmaps, redistribution $D=\text{micro}-\text{equal}$, and an overlay of the largest re-allocations on $x^1$.}}
\label{fig:mnist_allinone_maps}
\end{figure}

To explain $\Delta := g(\mathbf x_{\mathrm{end}})-g(\mathbf x^0)$, we compare the counterfactual equal-split Shapley (baseline \cite{luborgonovo2024}) and our micro-player Shapley attribution method.
In both cases, Shapley values are estimated by Monte-Carlo permutation sampling (to avoid enumerating $2^{|N|}$ subsets or $2^{|N'|}$ micro-coalitions). Pixel-level attribution maps are reported, such that totals match $\Delta$ by efficiency \cite{shapley1953}.
Fig.~\ref{fig:mnist_allinone_maps} first answers the counterfactual question ``what changes make the classifier see a $7$?'': the dominant positive attributions (blue) concentrate on the horizontal top stroke and the upper-right hook, \textit{i.e.}
the pixels that visually transform a thin ``1'' into a ``7''.
Negative contributions (red) appear on pixels that either preserve the ``1''-like vertical structure or add mass in regions that are inconsistent with a typical ``7'', thus, contributing to increase false positives.

\emph{Interaction explainability} (\cref{fig:mnist_allinone_maps}). The top-25 pixels largely overlap, showing broad agreement of the two attribution methods. The redistribution map reveals a localized reallocation of attribution mass, as micro-game Shapley (for $m=10$) amplifies a small subset of pixels and down-weights others (highlighted in the overlay, Figure~\ref{fig:mnist_allinone_maps} at the bottom),
reflecting interaction-aware responsibility along the counterfactual path rather than equal splitting (baseline Shapley).
Thus, the difference $D$ between methods is not \emph{whether} the counterfactual succeeds (both explain the same $\Delta$),
but \emph{where} responsibility is placed during transition.

\begin{figure}[t]
\centering
\includegraphics[width=\linewidth]{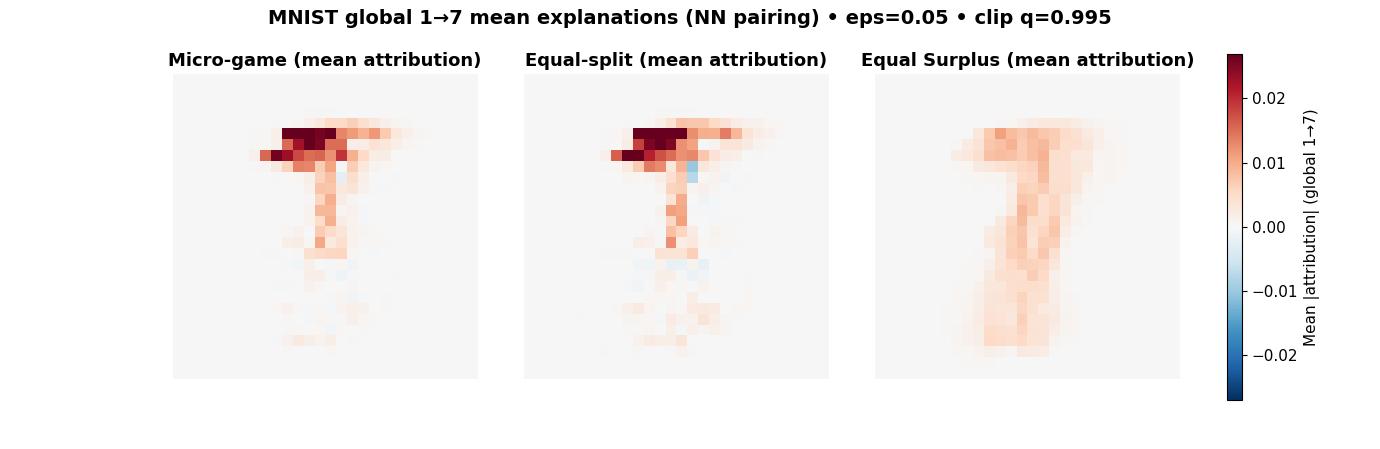}
\caption{
\scriptsize
\textbf{Global MNIST $1\to7$ mean explanations (nearest-neighbor pairing).
We sample $200$ test-set baselines $x^0$ of digit $1$ and pair each with its nearest neighbor $x^1$ among test-set $7$'s (Euclidean distance), then compute local attributions on the changed-pixel set $\{(r,c):|x^1_{rc}-x^0_{rc}|>\varepsilon\}$ with $\varepsilon=0.05$ and average the resulting maps over pairs.}}
\label{fig:mnist_global_mean_explanations}
\end{figure}

\emph{Global explainability} (\cref{fig:mnist_global_mean_explanations}). Explanations are aggregated across many $1\to7$ transitions and answer the question: \emph{which pixels are reliably responsible for preventing the classifier from producing a false positive ``1'' instead of a true positive ``7''?}.
Each attribution method concentrates on the characteristic ``7'' top bar and the upper hook region, indicating that, over the test distribution, these strokes are the most consistent drivers of the accuracy score. Concentration curves quantify \emph{how sharply} importance is localized: micro-game and equal-split Shapley concentrate a large fraction of the mean attribution mass into a small set of pixels, while Equal Surplus spreads mass more diffusely across the digit, producing a slower accumulation.

\noindent\begin{minipage}{\linewidth}
    \centering
    \includegraphics[width=\linewidth]{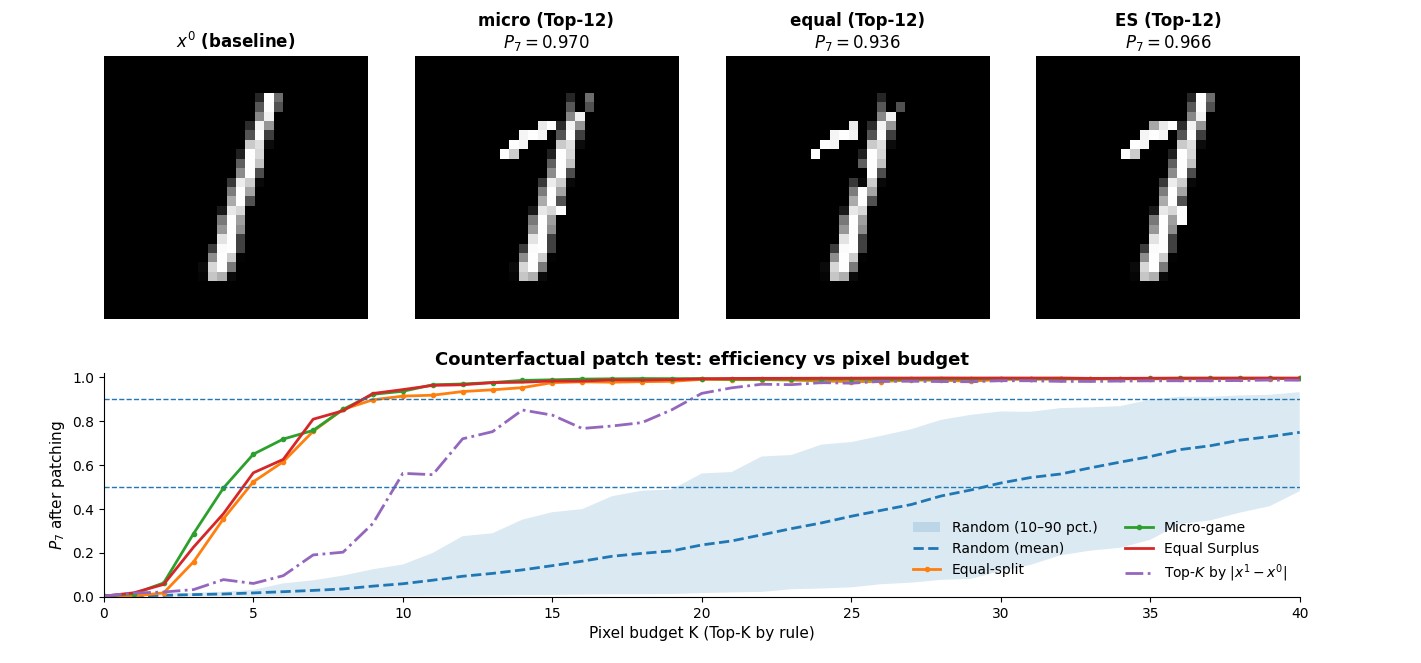}
    \captionof{figure}[Patchtest]{%
    \scriptsize
    \textbf{Counterfactual patch test on MNIST ($1\to7$): $P_7$ after patching the top-$K$ pixels selected by each rule (Micro-game Shapley, Equal-split Shapley, Equal surplus, and $|x^1-x^0|$), compared to a random baseline (mean with 10--90\% band). Top row shows $x^0$ and the patched images at $K=12$.}}
\label{fig:mnist_patch_budget_curve}
    \end{minipage}
    \smallskip
    
\emph{Counterfactual patch test} (\cref{fig:mnist_patch_budget_curve},
Table~\ref{tab:patch_metrics}).
Starting from baseline digit ``1'', we rank the 109 changed pixels
by each attribution rule and apply them sequentially toward ``7'',
recording $P_7$ at each budget $K$. Since all methods patch the
\emph{same} pixels and only the ordering differs, any gap is
attributable solely to the attribution rule.

Game-theoretic attribution vastly outperforms naive magnitude ordering: ranking by $|x^1-x^0|$ requires $K=14$ edits to reach $P_7\ge0.5$ versus $K=4$--$5$ for game-theoretic rules ($3.5\times$ fewer edits), and its AUC collapses to $0.583$ vs.\ $0.87$--$0.89$. Random ordering (mean of 200 trials, AUC$=0.109$) confirms the flip is not generic.

\begin{table}[H]
\centering
\caption{\scriptsize\textbf{MNIST patch test efficiency metrics.
Smaller $K@\tau$ and larger AUC are better.}}
\label{tab:patch_metrics}
\footnotesize
\setlength{\tabcolsep}{4pt}
\renewcommand{\arraystretch}{1.15}
\begin{tabular}{l c c c}
\toprule
\textbf{Method} & $K@0.5$ & $K@0.9$ & AUC\\
\midrule
Micro-game Shapley  & \textbf{4} & \textbf{6} & \textbf{0.8928}\\
Equal-split Shapley       & 5          & 6          & 0.8916\\
Equal Surplus             & 5          & 7          & 0.8761\\
$|x^1-x^0|$ magnitude    & 14         & 21         & 0.5823\\
Random (mean of 200)      & ---        & ---        & 0.1094\\
\bottomrule
\end{tabular}
\end{table}

Among game-theoretic methods, micro-game Shapley reaches $P_7\ge0.5$ in $K=4$ edits vs.\ $K=5$ for equal-split and ES, and $P_7\ge0.9$ in $K=6$ vs.\ $K=7$ for ES, achieving the highest AUC ($0.8928$). The gap over equal-split is small ($0.0012$) on this single transition, consistent with global agreement on top features; the advantage concentrates at small budgets ($K\le6$) where interaction geometry matters most.

\section{Discussion}\label{sec:discussion}

\noindent\emph{When does geometry matter?} Geometry-aware and equal-split attributions agree in most instances, but diverge where interactions are strong and asymmetric. In German Credit, rank flips occur in $12.3\%$ of instances, those with large Harsanyi pots where one feature enables the synergistic regime early. In MNIST, the advantage concentrates at small budgets ($K\le6$), where the first pixel changes determine whether downstream stroke interactions activate.

\noindent\emph{Actionability and causal constraints.} Aumann-SHAP takes any counterfactual pair $(\vx^0,\vx^1)$ as input; actionability constraints are enforced at the generation stage \cite{ustun2019actionable,karimi2021recourse}, which is standard in recourse literature. Within the pair, Aumann-SHAP explains which changed features enabled vs.\ harvested interactions. Integrating structural causal constraints directly into the slider cube is left for future work. SCMs \cite{pearl2009causality} require a known causal graph and do not decompose score changes into interaction pots along a continuous path; Aumann-SHAP is complementary.

\noindent\emph{Scalability and practical implementation.}
The grid-state formula costs $\mathcal{O}(k(m+1)^k)$ per
interaction pot, which is tractable for $k\le 6$ with
moderate $m$ (Table~\ref{tab:m_reco_models}). For larger
$k$, three strategies control cost without sacrificing
correctness. \textit{Interaction truncation:} retain only
pots with $|\phi_u|/|\Delta y| > \tau$ (we use $\tau=0.005$
in our experiments, with a fallback rate of $7.8\%$ on
German Credit). \textit{Monte Carlo sampling:} for $k>10$,
estimate Shapley values via random permutations of $N'_u$
(used for MNIST with $k=109$, $m=10$, 200 permutations).
\textit{Order-$K$ approximation:} restrict to pots with
$|u|\le K_{\max}$ and redistribute residuals via ES.
The $C^1$ smoothness assumption in
Theorem~\ref{thm:m_to_infty_diagonal_AS} covers GLMs and
MLPs; for tree ensembles the finite-grid attribution remains
valid and stable under the saturation rule, while the
integral limit is a smooth approximation. In our experiments,
the saturation rule (stop when max share change $<0.1$
percentage points across three consecutive refinements)
converged within $m=5$--$15$ for all tested models.

\section{Conclusion}\label{sec:conclusion}

\texttt{Aumann-SHAP} attributes score improvements
$\Delta y=g(\vx^1)-g(\vx^0)$ in a way that respects the geometry
of the counterfactual path, not just its endpoints. Proposition~\ref{prop:m1_equal_split}
recovers equal-split Shapley as the degenerate $m=1$ case;
Theorem~\ref{thm:m_to_infty_diagonal_AS} proves convergence to the diagonal Aumann--Shapley / IG
limit; and Proposition~\ref{prop:grid_state_closed_form} delivers polynomial grid-state computation.
On a synthetic benchmark with known ground truth, equal-split
carries a permanent $\pm0.167$ bias that no refinement removes,
while micro-game Shapley converges to the exact $1/3$--$2/3$
decomposition. On UCI Heart Disease, geometry awareness corrects
a sign error that equal-split cannot detect. On German Credit and MNIST, the method changes feature priorities precisely when interaction geometry affects recourse efficiency. These results confirm
that geometry-awareness is not cosmetic but structurally necessary.

\textit{Limitations and future work.} The grid-state complexity
is exponential in interaction order $k$; high-order interactions
require Monte Carlo sampling or truncation. Extending the framework
to causal counterfactuals with structural constraints, and to
settings where the generator is part of the optimization, are
natural next steps. We see particular promise in model auditing
and regulatory compliance, where the \emph{path} by which a
decision changes is as important as the decision itself.

\section*{Acknowledgments}
\textbf{The authors have no competing interests to declare.}

\bibliographystyle{IEEEtran}
\bibliography{references2}

\appendix
\section{Proofs}

\subsection*{Proof of Proposition~\ref{prop:ru_boundary_corner} (Properties)}

\noindent\textbf{(i) Boundary vanishing.}
Assume $t_{i_0}=0$ for some $i_0\in u$. Define the involution $\psi:\mathcal{P}(u)\to\mathcal{P}(u)$ by
\[
\psi(T):=
\begin{cases}
T\setminus\{i_0\}, & i_0\in T,\\
T\cup\{i_0\}, & i_0\notin T.
\end{cases}
\]
Then $|\psi(T)|=|T|\pm 1$, hence $(-1)^{k-|\psi(T)|}=-(-1)^{k-|T|}$.
Moreover, since $t_{i_0}=0$, the masked vectors coincide:
$
t^{(\psi(T))}=t^{(T)},
$
because whether $i_0\in T$ or not, the $i_0$-coordinate of the masked vector equals $0$.
Therefore the terms indexed by $T$ and $\psi(T)$ cancel pairwise in the sum defining $r_u(t)$, and thus $r_u(t)=0$.

\noindent\textbf{(ii) Far-corner equals the pot.}
Take $t=\one_k$. For each $T\subseteq u$, the masked vector $\one_k^{(T)}$ is the $0$--$1$ indicator of $T$, hence $\vx_u(\one_k^{(T)})=\vx^{T}$ and so
$g_u(\one_k^{(T)})=g(\vx^{T})$. Therefore
\[
r_u(\one_k)=\sum_{T\subseteq u}(-1)^{k-|T|}\,g(\vx^{T}).
\]
Using $g(\vx^{T})=V(T)+g(\vx^{0})$ yields
\begin{align*}
r_u(\one_k)
&=\sum_{T\subseteq u}(-1)^{k-|T|}\,V(T)
+
g(\vx^{0})\sum_{T\subseteq u}(-1)^{k-|T|}\\
&=\phi_u + g(\vx^{0})\,(1-1)^k
=\phi_u,
\end{align*}
since $\sum_{T\subseteq u}(-1)^{k-|T|}=(1-1)^k=0$.
\qedbox

\subsection*{Proof of Theorem~\ref{thm:micro_shapley} (Micro-player Shapley value)}

The induced map $v_u:2^{N'_u}\to\mathbb{R}$ is an ordinary TU-game on the finite player set $N'_u$.
Therefore, the classical Shapley characterization theorem applies directly: the unique value on games over $N'_u$
satisfying \textbf{E}, \textbf{S}, \textbf{A} and \textbf{N} is the Shapley value. Hence, for every micro-player
$(i,s)\in N'_u$, the displayed formula gives $\varphi^{\mathrm{Sh}}_{(i,s)}(v_u)$, and conversely the formula satisfies all four axioms.
\qedbox

\subsection*{Proof of Proposition~\ref{prop:m1_equal_split} (Equal-split Shapley)}

Assume $m_i=1$ for all $i\in u$. Then $N'_u=\bigcup_{i\in u}(\{i\}\times\{1\})$, $|N'_u|=|u|$, and for any $A\subseteq N'_u$, $p(A)\in\{0,1\}^k$, $v_u(A)=r_{p(A)}$.

\noindent\emph{Step 1: identify $r_p$ with $r_u(t(p))$ on the Boolean grid.}
For $p\in\{0,1\}^k$, $r_p=\sum_T(-1)^{k-|T|}g_{p^{(T)}}=r_u(t(p))$.

\noindent\emph{Step 2: proper coalitions have zero value.}
If $A\subsetneq N'_u$, then there is $i_0$ with $c_{i_0}(A)=0$, so $t(p(A))_{i_0}=0$, and by Boundary $r_u(t(p(A)))=0$.

\noindent\emph{Step 3: grand coalition equals the pot.}
$t(p(N'_u))=\one_k$, and by Efficiency $r_u(\one_k)=\phi_u$.

\noindent\emph{Step 4: Shapley splits a unanimity game equally.}
$v_u$ is the unanimity game on $N'_u$, so each micro-player gets $\phi_u/|u|$, and $S_i^{\mathrm{Sh}}(u)=\varphi^{\mathrm{Sh}}_{(i,1)}(v_u)=\phi_u/|u|$.
\qedbox

\subsection*{Proof of Theorem~\ref{thm:m_to_infty_diagonal_AS} (Aumann--Shapley limit)}

Fix $i\in u$ and write
$m_{\min}:=\min_j m_j$, $n:=|N'_u|=\sum_j m_j$.
Let $\Pi$ be a uniform random permutation of $N'_u$ and $A_t:=\{\Pi_1,\dots,\Pi_t\}$.

\noindent\emph{Step 1.} Random-order Shapley representation:
\begin{multline*}
S^{\mathrm{Sh}}_{i\to u}=\E\Big[\sum_{t=0}^{n-1}\mathbf{1}\{\Pi_{t+1}\in\{i\}\times\{1,\dots,m_i\}\}\\
\times(v_u(A_t\cup\{\Pi_{t+1}\})-v_u(A_t))\Big].
\end{multline*}

\noindent\emph{Step 2.} Marginals as one-step residual differences:
on the event $\{\Pi_{t+1}=(i,s)\}$, $p(A_t\cup\{(i,s)\})=p(A_t)+e_i$, so the marginal equals $r_u(t(p(A_t))+\tfrac{1}{m_i}e_i)-r_u(t(p(A_t)))$.

\noindent\emph{Step 3.} Mean-value: $r_u\in C^1$ gives $\theta_t\in(0,1)$ with the difference equal to $\tfrac{1}{m_i}\partial_i r_u(t(p(A_t))+\tfrac{\theta_t}{m_i}e_i)$.

\noindent\emph{Step 4.} For each $j$, with $C_j(t)=c_j(A_t)$, the Dvoretzky--Kiefer--Wolfowitz inequality for sampling without replacement gives
\[
\PP\Big(\max_t\big|C_j(t)/m_j-t/n\big|\ge\varepsilon\Big)\le 2e^{-2m_j\varepsilon^2},
\]
so $t(p(A_t))$ concentrates on the diagonal $(t/n)\one_k$.

\noindent\emph{Step 5.} Replace evaluation points by the diagonal using uniform continuity of $\partial_i r_u$.

\noindent\emph{Step 6.} By symmetry, $\PP(\Pi_{t+1}\in\{i\}\times\{1,\dots,m_i\})=m_i/n$, so
\[
S^{\mathrm{Sh}}_{i\to u}=\frac{1}{n}\sum_{t=0}^{n-1}\partial_i r_u\big(\tfrac{t}{n}\one_k\big)+o(1).
\]
The Riemann sum converges to $\int_0^1\partial_i r_u(\tau\one_k)\,d\tau$ as claimed.

\noindent\emph{Efficiency of the limit.} By the chain rule and Boundary/Efficiency of $r_u$,
$\sum_{i\in u}\int_0^1\partial_i r_u(\tau\one_k)d\tau = r_u(\one_k)-r_u(\zero)=\phi_u$.
\qedbox

\subsection*{Proof of Proposition~\ref{prop:grid_state_closed_form} (LES closed form)}

Recall $v_u(A)=r_{p(A)}$ and $|A|=|p(A)|$. For fixed $(i,s)\in N'_u$,
\begin{multline*}
\varphi^{\mathrm{LES}}_{(i,s)}(v_u)=
\sum_{A\subseteq N'_u\setminus\{(i,s)\}}\frac{|A|!(n-|A|-1)!}{n!}\\
\times[b(|p(A)|+1)r_{p(A)+e_i}-b(|p(A)|)r_{p(A)}].
\end{multline*}
Group by grid state $p$: the number of $A\subseteq N'_u\setminus\{(i,s)\}$ with $p(A)=p$ is $\binom{m_i-1}{p_i}\prod_{j\ne i}\binom{m_j}{p_j}$.
Summing over the $m_i$ replicas of feature $i$ multiplies by $m_i$, and using $m_i\binom{m_i-1}{p_i}=(m_i-p_i)\binom{m_i}{p_i}$,
\[
S^{\mathrm{LES}}_{i\to u}=\sum_{\substack{p:\,p_i<m_i}}\frac{|p|!(n-|p|-1)!}{n!}\prod_j\binom{m_j}{p_j}(m_i-p_i)\Delta_p(i).
\]
\qedbox
\end{document}